\documentclass[10pt,twocolumn,letterpaper]{article}

\usepackage{cvpr}
\usepackage{times}
\usepackage{epsfig}
\usepackage{graphicx}
\usepackage{amsmath}
\usepackage{amssymb}
\usepackage{multirow}
\usepackage{multicol}
\usepackage{booktabs}
\usepackage[hyphens,spaces,obeyspaces]{url}
\usepackage{flushend}

\usepackage[pagebackref=true,breaklinks=true,letterpaper=true,colorlinks,bookmarks=false]{hyperref}

\cvprfinalcopy % *** Uncomment this line for the final submission

\def\cvprPaperID{2261} % *** Enter the CVPR Paper ID here

\ifcvprfinal\fi

\begin{document}

%%%%%%%%% TITLE
\title{Who's Better? Who's Best? Pairwise Deep Ranking for Skill Determination}% with Pairwise Deep Ranking}

\author{Hazel Doughty \qquad Dima Damen \qquad Walterio Mayol-Cuevas\\
University of Bristol, Bristol, UK\\
{\tt\small <Firstname>.<Surname>@bristol.ac.uk}
}
\maketitle
\thispagestyle{empty}

%%%%%%%%% ABSTRACT
\begin{abstract}
\vspace{-0.5em}
This paper presents a method for assessing skill from video, applicable to a variety of tasks, ranging from surgery to drawing and rolling pizza dough. We formulate the problem as pairwise \textit{(who's better?)} and overall \textit{(who's best?)} ranking of video collections, using supervised deep ranking.
We propose a novel loss function that learns discriminative features when a pair of videos exhibit variance in skill, and learns shared features when a pair of videos exhibit comparable skill levels.
%to learn discriminative features between pairs of videos exhibiting \Dima{variance in}
%different amounts of 
%skill \Dima{when presence and the }
%. 
%We extend this deep ranking model to incorporate a new loss to learn similarities between comparably ranked videos. %utilise a two-stream Temporal Segment Network to capture both the type and quality of motions and the evolving task state. 
Results demonstrate our method is applicable across tasks, with the percentage of correctly ordered pairs of videos ranging from 70\% to 83\% for four datasets.
We demonstrate the robustness of our approach via sensitivity analysis of its parameters. 

We see this work as effort toward the automated organization of how-to video collections and overall, generic skill determination in video.
\end{abstract}

%%%%%%%%% BODY TEXT
\vspace{-0.8em}
\section{Introduction}
How-to videos on sites such as YouTube and Vimeo, have enabled millions to learn new skills by observing others more skilled at the task. From drawing to cooking and repairing household items, learning from videos is nowadays a commonplace activity. However, these loosely organized collections normally contain a mixture of contributors with different levels of expertise. The querying person needs to decide who is better and who to learn from. 
%While popularity scores can sometimes help, these are prone to subjective ratings and worse, cheating. 
Furthermore, the number of  \textit{how-to} videos is only likely to increase, fueled by more cameras recording our daily lives. An intelligent agent that is able to assess the skill of the subject, or rank the videos based on the skill displayed, would enable us to delve into the wealth of this on-line resource.  

In this work, we attempt to determine skill for a variety of tasks from their video recordings. We base this work on two assumptions, first - for tasks where human observers {\it consistently} label one video as displaying \textit{more skill} than another, there is enough information in the visual signal to automate that decision; and second - the same framework for determining skill can be used for a variety of tasks ranging from surgery to drawing and rolling pizza dough.

\begin{figure}[t]
\begin{center}
\includegraphics[width=\linewidth]{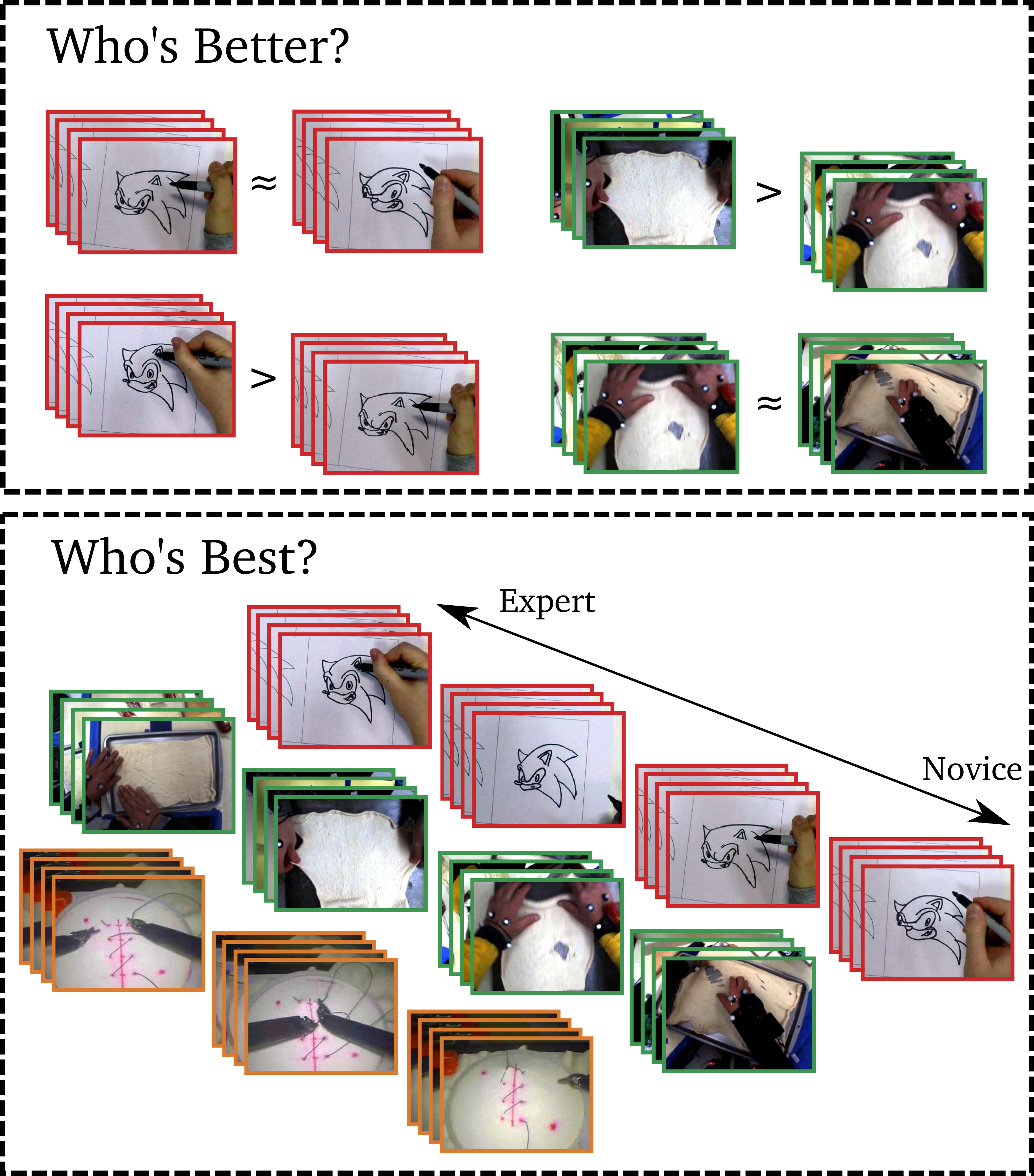}
\end{center}
\vspace{-2.5mm}
\caption{Determining skill in video. \textbf{Who's Better?} (Top): pairwise decisions of videos containing the same task, performed with varying or comparable levels of skill. \textbf{Who's Best?} (Bottom): ranking learned from pairwise decisions.}
\label{fig:concept}
\vspace{-3mm}
\end{figure}

We propose to determine skill using a pairwise deep ranking model, which characterizes the difference in skill displayed between a pair of videos, where one is ranked higher than the other \textit{by human annotators}~(Fig~\ref{fig:concept}). 
%also include a new pairwise similarity loss, which is used to learn a representation such that similarly ranked videos are indistinguishable. 
We use a Siamese architecture where \textit{each stream} is made up of a two-stream (spatial and temporal) convolutional neural network (2S-CNN). %composed of a spatial and temporal stream. 
This Siamese architecture is trained using a novel ranking loss function that considers the extent of the task within the video, and includes pairs of videos where the skill level is indistinguishable.
By assigning videos a relative score of skill for the given task, we can predict a \textit{skill ranking} for a set of videos. 
% * <wmayol@gmail.com> 2017-11-14T00:16:52.156Z:
% 
% > - in line with previous methods
% I feel we can remove this as it implies other works have done exactly what we have done.
% 
% ^ <wmayol@gmail.com> 2017-11-14T00:17:29.964Z.

Our main contributions are as follows: i)~We present the first method to determine skill in videos for a wide variety of tasks. ii) We propose a novel ranking loss function which considers the extent of the video and incorporates pairwise similarities in training. This loss function outperforms the standard ranking loss on all datasets by up to 5\%.
%augment the existing pairwise ranking framework with a new pairwise similarity loss. 
iii)~We present pairwise skill annotations for three datasets, two of which are newly recorded.
iv)~We evaluate our approach on four datasets (two public); one surgical - for which there is authoritative expert ranking, another on rolling pizza dough, as well as two newly introduced datasets for the tasks of drawing and using chopsticks. 
Newly recorded datasets and annotations are available from the authors' webpages. %or from: \url{http://github.com/hazeld/WhosBetterWhosBest}.}
%\Dima{Except for surgery, where annotations are provided,} we crowd-source the ground truth. Our proposed method 
%outperforms the baselines on three out of four datasets, 
%\DimaN{[the baseline isn't something we are interested in, as there's no baseline for this problem, so mentioning it here isn't helpful]}
%\Dima{achieves} a correctly ordered video pair accuracy of 72.5\% in the surgical tasks, 79.5\% in the dough rolling task, 83.2\% for the drawing tasks and 70.8\% in the chopstick-using task.

\section{Related Work}
In this section, we review skill determination works in video, primarily for surgical tasks and within sports. We relate this work to the new surge for utilizing collections of \textit{how-to} instructional videos. Finally, we introduce deep ranking approaches, on which our method is based.

\medskip
\noindent \textbf{Skill Determination.} There have been few prior works on automatically determining skill from video. The majority of these works are focused on surgical tasks ~\cite{malpani2014pairwise, sharma2014video, sharma2014automated, zhang2011video, zhang2015relative, zia2016automated,  zia2015automated,  zia2017video}, due to the intensive training needs in this area. For instance, Sharma \etal.~\cite{sharma2014video} use motion textures to predict the OSATS criteria: a measure of skill specific to the surgical domain. In~\cite{zia2016automated}, Zia \etal rely on the repetitive nature of surgical tasks, using the entropy of repeated motions to identify different skill levels. 
Malpani~\etal~\cite{malpani2014pairwise} use a combination of video and kinematic data to rank performance in two surgical tasks. However, they decompose each task into a sequence of actions, and design specific features for performance evaluation of surgical maneuverer, which makes this inapplicable to non-surgical tasks.
Generally, the high specialty of the tasks and methods involved in surgery make these approaches difficult to generalize.% outside the surgical domain. 

Many of these methods take a coarse %\Dima{video-level} \DimaN{[not sure we can call ours fine-grained]} 
%fine-grained 
approach to identifying skill, splitting participants into categories of novice and expert~\cite{zia2015automated}. Often skill labels are determined
%this is done 
by participants' previous experience, instead of their performance in individual videos~\cite{gao2014jhu,zia2015automated}. We aim however, to rank the performance in each video, instead of classifying the video, or all of a participant's videos, as expert or novice. 
  
A work that utilizes ranking for surgical tasks is that of Zhang \etal~\cite{zhang2015relative}. It uses relative Hidden Markov Models to evaluate human motion skill by obtaining a ranking between %input 
pairs. 
%\DimaN{[I'm confused about this sentence, why do you call then `input' pairs? As that confuses training as well. You can get rid of input?]}
%However, the main focus in this paper is again motion skill in surgical training tasks. 
This work is somewhat limited by the ground truth data: the assumption is that a video recorded at a later date will capture better performances than a participant's earlier recording. Thus, skill is only compared
within a participant's performances. 

There is also some skill assessment work in the domain of sport~\cite{bertasius2016baller, cceliktutan2013graph,ilg2003estimation,jug2003trajectory,parisi2016human,parmar2016learning, pirsiavash2014assessing}. However, many of these works are not generalizable to domains outside sports as they either craft features specific to a sport, such as basketball~\cite{bertasius2016baller,jug2003trajectory}, or focus on quality of motion~\cite{cceliktutan2013graph,ilg2003estimation,parisi2016human,parmar2016learning}. The most relevant of these works is from Pirsiavash \etal~\cite{pirsiavash2014assessing}, who present a general method for assessing the quality of actions. This is done by estimating human body pose with a skeleton model in order to predict the score of actions, again in sports videos. However, quality of motion on its own is not an essential condition to determine skill. For example, moving a brush in an artistic manner is not a sufficient measure for painting skills.

\medskip
\noindent \textbf{How-To Videos.} Related to skill determination are works on instructional videos~\cite{Alayrac16unsupervised, Damen2016_CVIU, Alayrac_2017_ICCV,kim2017evaluationnet},
that study
%in which there are a large amount of 
videos of different people performing the same activity. However, none of these works determine skill from these videos, focusing instead on aligning the steps undertaken to complete the task~\cite{Alayrac16unsupervised} and the object states and manipulations that occur during the task~\cite{Alayrac_2017_ICCV}. Kim et al.~\cite{kim2017evaluationnet} %come closest to evaluating skill by evaluating 
evaluate the semantic similarity of action units to determine if two people are performing the same sub-activity, however this is not capable of assessing the skill within the same task or sub-tasks.%ranking the skill shown in different videos within the same task.

%\medskip
%\noindent \textbf{Video Representation.} Recent approaches utilising deep learning to extract features, particularly CNNs, have shown great success in tasks, including image classification~\cite{krizhevsky2012imagenet}, object detection~\cite{girshick2014rich} and action recognition~\cite{simonyan2014two}. 
%A common consideration for these works is how to represent the video as an input to the network. Many mainstream CNNs~\cite{simonyan2014two,tran2015learning} focus on appearances and short term motions, ignoring any long range temporal structure.

%In this paper, we utilise the recent Temporal Segment Networks (TSN) architecture~\cite{wang2016temporal}, to model long range temporal structure. This architecture achieves state of the art performances for action recognition on UCF101~\cite{soomro2012ucf101} and HMDB51~\cite{kuehne2011hmdb}. TSN decomposes the input video into uniformly sized segments, sampling a snippet from each segments as input to a CNN. The long range temporal structure is modelled by forming a consensus from these snippets, before the result is fed into the loss layer during training. This method has the advantage of enabling end-to-end learning for long video sequences with a relatively low cost in terms of time and computing resources.
%\Hazel{I feel like this section isn't particularly important, may get rid of it and add couple of sentences for motivation for using TSN elsewhere}

\begin{figure*}
\begin{center}
\includegraphics[width=\textwidth]{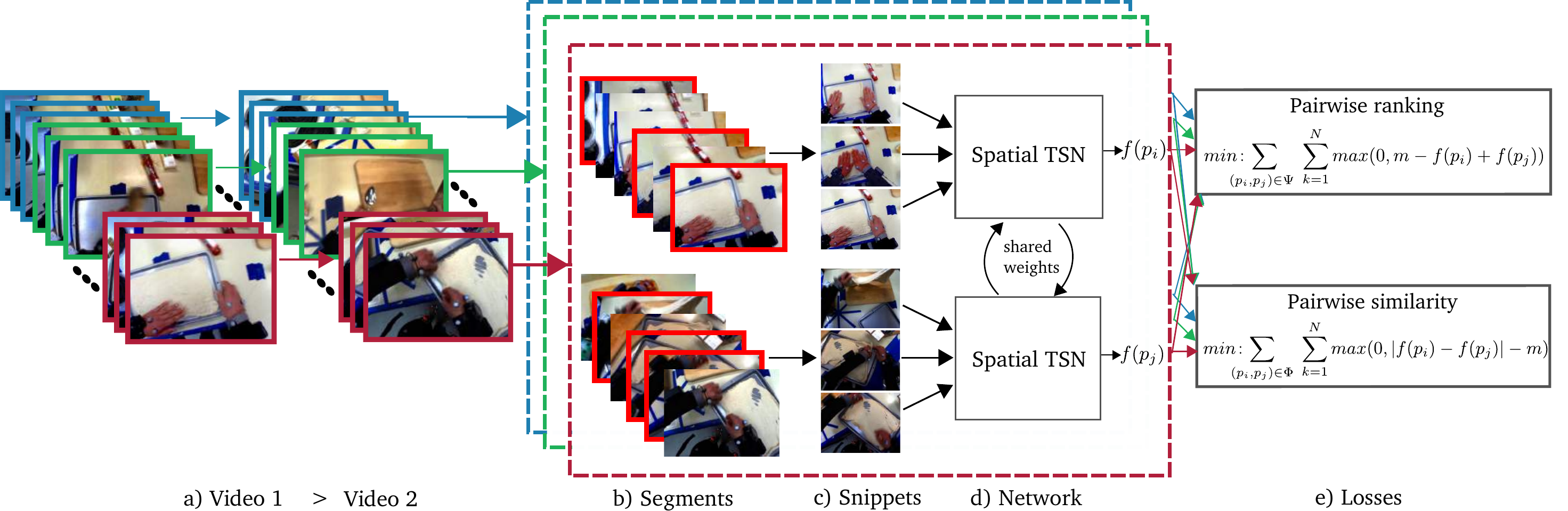}
\end{center}
\vspace{-1mm}
	\caption{Training for skill determination. a) We consider all pairs of videos, where the first is showing a higher level of skill $\Psi$, or their skill is comparable $\Phi$, and divide these into $N$ splits to make use of the entire video sequence. b) Paired splits are then divided up into 3 equally sized paired segments as in~\cite{wang2016temporal}. c) TSN selects a snippet randomly from each segment. For the spatial network this is a single frame, for the temporal network this is a stack of 5 dense horizontal and vertical flow frames. d) Each snippet is fed into a Siamese architecture of shared weights, for both spatial and temporal streams, of which only the spatial is shown here. e) The score from each split is either fed to the proposed loss functions: ranking/similarity which compute the margin ranking loss based on the pair's label.}
    \vspace{-1mm}
    \label{fig:network}
\end{figure*}

\medskip
\noindent \textbf{Deep Ranking.} The most widely used learning to rank formulation is pairwise ranking. The method aims to minimize the average number of incorrectly ordered pairs of elements in a ranking, by training a binary classifier to decide which element in a pair should be ranked higher. This formulation was used by Joachims in RankSVM~\cite{joachims2002optimizing}, where a linear SVM is used to learn a ranking. %the standard SVM algorithm 
%\DimaN{[There's nothing called an SVM algorithm]}
%is adapted to solve ranking problems via learning to rank. 
It was originally used to learn search engine retrieval functions from click-through data, however it has been adopted in other ranking applications, including ranking relative attributes in images~\cite{parikh2011relative}.

Pairwise ranking has also been used in deep learning, 
%in deep learning with gradient descent
first by Burges et al.~\cite{burges2005learning} with RankNet. 
%Other methods of deep ranking have since been developed, such as LambdaMART~\cite{burges2010ranknet} which incorporates a list-wise approach, however the pairwise approach based on RankNet is the most commonly used and the one we adopt. 
For instance, Yao \etal~\cite{yao2016highlight} use a pairwise deep ranking model to perform highlight detection in egocentric videos using pairs of highlight and non-highlight segments. They use a ranking form of hinge loss as opposed to the binary cross entropy loss used in RankNet. In our paper we base our ranking loss on the pairwise margin loss used by Yao \etal, but with several novel additions, including a pairwise similarity loss. 

Other non-pairwise methods for deep ranking such as list-wise ranking~\cite{burges2010ranknet} have been proposed, yet are less frequently used compared to the pairwise approach, unless optimizing for a specific evaluation metric such as NDCG~\cite{jarvelin2000ir}. %\DimaN{[due to? difficulty in optimisation? inferior performance?]}

\section{Learning to Determine Skill}
\label{sec:method}

In this section we first give an overview of the skill determination problem and the Siamese two-stream CNN architecture we use to determine skill. We then present our novel additions to the pairwise margin loss function used to train both streams of the CNN.

\subsection{Problem Definition}
\label{sec:def}
Our goal is to learn models for ranking skill in different tasks.
Given a task, we have a set of $K$ videos $P = \{p_k, 1 \le k \le K\}$, from multiple people, each performing the task one or more times.
We consider each video independently, 
%even if performed by the same person.
%We 
%and assume that 
as people differ in the skill they display in each video, even across multiple runs.
We are thus interested in ranking relative skill per video instead of accumulating a score per person.

\begin{equation}
E(p_i, p_j) = 
\begin{cases}
1 &p_i \text{ shows higher skill than } p_j\\
-1 &p_j \text{ shows higher skill than } p_i\\
0 &\text{no skill preference}
\end{cases}
\label{eq:annotationsExpertise}
\end{equation} 
Note that according to Eq.~\ref{eq:annotationsExpertise}, $E(p_i, p_j) = -E(p_j, p_i)$, we thus need to only obtain one annotation for each pair. We explain how these annotations are obtained in Section~\ref{sec:annotation}.

\subsection{Time as a Measure of Skill}

A naive way to approach measuring skill is to use time of completion, as finishing a task faster (or slower) could imply a higher level of skill. However, from the JIGSAWS dataset~\cite{gao2014jhu} we prove that time is not sufficient. Although there is some correlation between score and time in the Knot Tying task ($\rho = 0.72$), there is little correlation in the Needle Passing ($\rho=0.23$) and Suturing ($\rho=0.34$) tasks. Therefore, although time can be useful in some tasks, it is not a general or reliable method for skill determination. We thus propose a method for skill determination that is independent of time of completion. %and thus our method for skill determination is independent of time of completion.

\subsection{Temporal Segment Networks as Architecture}

\label{sec:two-stream}

Tasks differ in how skill can be demonstrated. In this respect we identify two main sources of relevant information. The first is the quality and type of motions used. The second is the effect on the environment captured through the appearance of the task. We thus utilize two stream convolutional neural networks (2S-CNN) for skill determination. Specifically, we base our method on Temporal Segment Networks (TSN)~\cite{wang2016temporal}. We select TSN due to their state of the art performances on action recognition benchmarks and ability to model long range temporal structure and dependencies. %This makes the approach suitable for determining skill for tasks \Dima{for both short and long } regardless of the length of the videos used.

In training TSN, as in~\cite{wang2016temporal}, we uniformly divide each input video sequence into three segments, then randomly sample a single short snippet from each of these segments (Fig~\ref{fig:network}b,c). For each iteration in training, our 2S-CNN outputs a preliminary prediction of skill for each snippet. This decision is then pooled across the three snippets, creating a score per input video. The output to the loss function (Fig.~\ref{fig:network}e), in both the spatial and temporal streams, is then the consensus between selected snippets. 
\subsection{Pairwise Deep Ranking}

\label{sec:training_pairwise}
%As we want to determine relative skill of users in different videos, 
We use the pairwise approach for learning to rank. To do this we build a Siamese version of the two-stream TSN described in Section~\ref{sec:two-stream}, with the weights shared across both sides of the Siamese network (Fig.~\ref{fig:network}d). Given a pair of videos, where the first video is ranked higher than the second in terms of skill, we want the Siamese network to output a higher score for the first. Formally, we have a set of pairs $\Psi = \{(p_i, p_j); E(p_i, p_j)=1\}$ (ref Eq.~\ref{eq:annotationsExpertise}).
These two videos are fed into the separate, but identical, TSNs which form the Siamese network (Fig~\ref{fig:network}a). Assuming the TSN outputs $f(\cdot)$, our goal is to learn the function $f$ such that we determine skill, where 
\begin{equation} 
f(p_i) > f(p_j) \quad \forall (p_i, p_j) \in \Psi
\end{equation}

To gain an overall rank for all videos, we use a margin loss layer to evaluate the loss for each pair. The loss function we use is an approximation to 0-1 ranking error loss that has been used successfully for other applications~\cite{yao2016highlight, wang2014learning};
\vspace{-1mm}
\begin{equation}
\label{eq:loss}
L_{rank1} = \sum_{(p_i, p_j) \in \Psi} max(0, m - f(p_i) + f(p_j))
\end{equation}

We use $m=1$ in our experiments. During training, this loss function evaluates the violation of the ranking of each pair of videos and back-propagates the gradient through the network. This allows the network to learn discriminative features to distinguish between the amount of skill displayed in different videos.

\subsection{Pairwise Deep Ranking with Splits}
\label{sec:dataaug}
Traditionally, 2S-CNN are used for action recognition~\cite{simonyan2014two}, thus the whole length of the video needs to be considered once to recognize the undertaken action. In this work, we are examining skill, which could be understood from all (or any) parts of the video sequence. To make the most of the extent of the video sequence, we 
%change our network to split videos into 
consider $N$ uniform splits (Fig.~\ref{fig:network}a) and evaluate each of the corresponding splits in the loss function. We assume that two videos of the same task have comparable rate of progression through the task, and thus compare the temporal splits across a pair of videos in order. Assume $p_i^k$ is the $k^{th}$ split of video $p_i$, we extend the skill annotations such that,
\begin{equation}
E(p_i^k, p_j^k) = E(p_i, p_j) \quad \forall k = 1 \cdots N
\end{equation}
Our loss function now becomes:
\vspace{-1mm}
\begin{equation}
\label{eq:data_aug_loss}
L_{rank2} = \sum_{(p_i, p_j) \in \Psi} \ \sum_{k=1}^N max(0, m - f_k(p_i) + f_k(p_j))
\end{equation}
\vspace{-1mm}

\noindent In our experiments, $N = 7$ was tested.
By pairing corresponding splits, we ensure the two videos are compared at a similar stage of the task performance, while still being able to deal with videos of different lengths, and therefore more discriminative features are likely to be learned.

\subsection{Pairwise Deep Ranking with Similarity Loss}
\label{sec:sim_loss}
With the margin loss function in Section~\ref{sec:training_pairwise} we only incorporate pairs where one video is consistently ranked higher than another. In order to utilize more of the potential video pairings, we take inspiration from recent works in domain adaptation~\cite{ganin2016domain} by creating a secondary `adversarial' loss where we wish to not distinguish between our similar pairs. 
%Here we want something similar. 
We modify the margin loss to learn features which map pairs, indistinguishable in terms of skill, to similar scores. 
We thus find the set of pairs with indistinguishable skill levels
$\Phi = \{(p_i, p_j); E(p_i, p_j)=0\}$ (ref Eq.~\ref{eq:annotationsExpertise}).

The way in which adversarial loss function are commonly created is by reversing the gradient, however  this does not work in a ranking problem. In order to learn indistinguishable representations we aim for the following:
\vspace{-5mm}
%therefore instead of learning how to distinguish between classes the network instead learns a representation in which they cannot be distinguished. However, if we reverse the gradient of Equation~\ref{eq:loss} the network will instead learn an opposite ordering of pairs. In this case, instead the previous aim \DimaN{[what do you mean by instead the previous aim]}:

% \begin{equation}
% f(p_i) \geq f(p_j) + m \equiv f(p_j) - f(p_i) + m \leq 0
% \vspace{-1mm}
% \end{equation}
% We now want:
% \vspace{-1.5mm}
\begin{equation}
|f(p_i) - f(p_j)| \leq m \equiv |f(p_j) - f(p_i)| - m \leq 0
\end{equation}
Therefore, our new loss function for similar pairs becomes:
\begin{equation}
L_{sim} = \sum_{(p_i, p_j) \in \Phi} \ \sum_{k=1}^{N} max(0, |f(p_i) - f(p_j)| - m)
\label{eq:l_sim}
\end{equation}
Resulting in a modified loss function:
\vspace{-1mm}
\begin{equation}
L_{rank3} =  \beta L_{rank2} + (1-\beta) L_{sim}
\label{eq:finalLoss}
\end{equation}
\vspace{-1mm}

\noindent Adding $L_{sim}$ into our ranking loss for similarly ranked pairs not only allows us to utilize extra data pairs in the learning process, but also encourages the network to learn similarities in skill between similarly ranked videos. We explain how we get the new set of pairs $\Phi$ in Section~\ref{sec:annotation}. %The value of $\beta$ is chosen such that the difference between the amount of pairs to rank and similar pairs is balanced out during training.

\subsection{Evaluating Skill for a Test Video}
\label{sec:testing}
Following training, the learned 2S-CNN weights are used to evaluate the skill for test videos of the same task. In testing, we uniformly sample $\sigma$ snippets from each video $p_i$, again as in~\cite{wang2016temporal}. Each snippet $p_{i}^k\; 1 \le k \le \sigma$ is then fed into the spatial and temporal TSN independently. The output for each snippet is a score $f(p_{i}^k)$ for both spatial $f_s(p_{i}^k)$ and temporal $f_t(p_{i}^k)$ streams. To fuse the spatial and temporal networks for all snippets we take the weighted average of the outputs,
\vspace{-2mm}
\begin{equation}
f(p_i) = \frac{1}{\sigma} \sum_{k=1}^\sigma \alpha f_s(p_{i}^k) + (1-\alpha) f_t(p_{i}^k)
\label{eq:alpha}
\end{equation}
where $\alpha$ is the fusion weighting between spatial and temporal information, and $\sigma$ is the number of testing snippets.

An overall ranking for a test set is achieved by ordering all test videos in a descending order based on $f(p_i)$.

\begin{figure}
\begin{center}
	\includegraphics[width=0.97\linewidth]{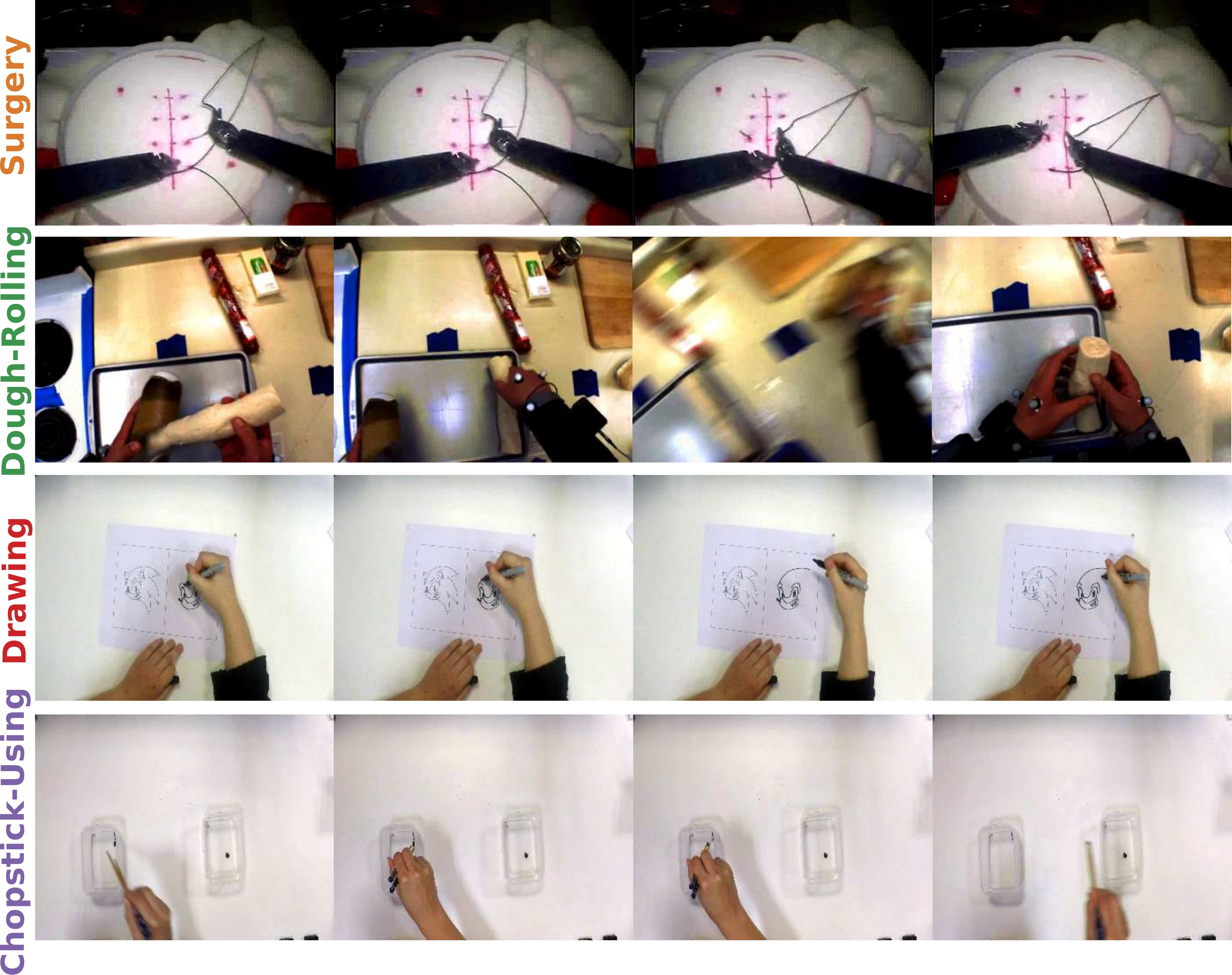}
\end{center}
\vspace{-4mm}
\caption{Sample sequences from the four tasks.}
\label{fig:datasets}
\vspace{-2.5mm}
\end{figure}

\section{Tasks and Datasets}
For evaluation we conduct experiments on tasks from four datasets - two published and two newly recorded (Fig.~\ref{fig:datasets}). The first is a surgical dataset.
Three other datasets containing daily living tasks are also used, to demonstrate the generality of the approach. Here we detail the four datasets, followed by the skill annotations for these datasets. These datasets and annotations will be combined to form the new EPIC-Skills 2018 dataset which can be found on the authors' webpages. %at: \url{http://github.com/hazeld/WhosBetterWhosBest}.

\medskip
\noindent \textbf{Surgery.} We use the published JHU-ISI Gesture and Skill Assessment Working Set (JIGSAWS) dataset~\cite{gao2014jhu}. In this dataset, three surgical procedures are performed by 8 surgeons with varying levels of experience. In total, JIGSAWS consists of 36 trials of Knot Tying, 28 trials of Needle Passing and 39 trials of Suturing. This dataset contains stereo recordings, from which we use only one video (right view) from each sequence.

\medskip
\noindent \textbf{Dough-Rolling.} We use the kitchen-based CMU-MMAC dataset~\cite{de2008guide}, and select the dough rolling task from the pizza making activity, as this exhibits varying levels of performance across participants. In total, we manually segment 33 Dough-Rolling videos from 33 distinct participants.

\medskip
\noindent \textbf{Drawing.} We introduce a new dataset for drawing, captured using a stationary camera at a resolution of 1920x1080 and a frame rate of 60 fps. Participants were given a reference image to copy. Two reference images were used; a cartoon of Sonic the Hedgehog and a gray-scale photograph of a hand. Similarly to the \textbf{Surgery} tasks, both tasks were performed five times each, by four participants. 

\medskip
\noindent \textbf{Chopstick-Using.} We also introduce a new dataset for using chopsticks, captured using the same setup as the \textbf{Drawing} dataset. 
%Participants were presented with two identical tubs, one containing four coffee beans. 
Each participant was tasked with moving as many of the beans as possible from one tub to the other using chopsticks, limited to one minute per trial. Eight participants were recruited, each repeated the task five times.% and were limited to one minute per trial. 

\subsection{Skill Annotation}
\label{sec:annotation}
Only the JIGSAWS dataset has existing skill scores. This was annotated by a surgery expert, out of a maximum score of 30. In this section, we explain how we obtained skill ranking for the remaining three datasets using \textit{Amazon Mechanical Turk (AMT)}.

We determine the ground truth relative ranking of video pairs using a similar method to~\cite{malpani2015study}, where the authors demonstrate crowdsourcing yields reliable pairwise comparison for skill in surgical tasks. We asked AMT workers to watch pairs of videos simultaneously and select the video displaying the higher level of skill for the given task. Each worker was presented with 5 pairs of videos per HIT from the same task, one of which was a quality control pair which displayed an obvious difference in skill. Annotators were asked for strict preferences per pair. We then check for consensus between different annotators for skill annotation. Each video pair was annotated by four different workers. Only pairs of videos for which \textit{all} annotators agreed on their skill order are considered for training in the $L_{rank}$ loss function, we refer to these as \textit{consistent pairs}.

We further check for any discrepancies in the set of these \textit{consistent pairs}, by checking for \textit{triangular inconsistencies}. Assume ${E(p_i, p_j) > 0}$ and ${E(p_j, p_k) > 0}$, we check for $E(p_i, p_k) < 0$, which would show a triangular inconsistency in annotations. We do this by creating a directed graph with $P$ nodes and edges ${(p_i \rightarrow p_j) \: \forall \, 1 \le i,j \le K \text{ where } E(p_i,p_j) > 0}$. Cycles in the graph would indicate a triangular inconsistency, which we manually resolve. 
%For instance the ranking of videos $p_i$, $p_j$ and $p_k$ would be inconsistent if the set of ordered pairs contained the pairs $p_i > p_j$, $p_j > p_k$ and $p_k > p_i$. 
Only a single triangular inconsistency was found in all AMT annotations for the three tasks. This was in the \textbf{Dough-Rolling} task and was excluded from training and testing. Similarly, we take the skill scores from the \textbf{Surgery} dataset and compute all \textit{consistent pairs}. 

\begin{table}[t]
\begin{center}
\begin{tabular}{@{}c@{\:}c@{\:}c@{\:}c@{\:}c@{}c@{\:}}
	\hline \rule{0pt}{0.4\normalbaselineskip}
    \multirow{2}{*}{Task} &\#Vid- & \#Max &\%Cons. & \%Sim. & Total\\ 
    &eos &Pairs &Pairs &Pairs & Pairs\\ \hline \rule{0pt}{0.4\normalbaselineskip}
    Surgery (KT) & 36 & 630  &95\% & 5\% & 100\%\\
    Surgery (NP) & 28 &378  &96\% & 4\% & 100\%\\
    Surgery (Suturing) & 39  &701 &95\% & 5\% & 100\%\\ \hline \rule{0pt}{0.4\normalbaselineskip}
    Dough-Rolling & 33 &528  &34\% & 18\% & 52\%\\ \hline \rule{0pt}{0.4\normalbaselineskip}
    Drawing (Sonic) & 20 &190  &62\% & 37\% & 99\%\\ 
    Drawing (Hand) & 20 & 190  &68\% & 26\% & 94\%\\ \hline \rule{0pt}{0.4\normalbaselineskip}
    Chopstick-Using & 40 &780  &69\% & 10\% & 79\%\\
    \hline
\end{tabular}
\end{center}

\caption{For the four datasets: \#videos, \#of pairs $(n)(n-1)/2$ with the percentage of consistent pairs in annotations and similar pairs obtained. KT=Knot Tying, NP=Needle Passing}
\label{tab:pairs}
\vspace{-2mm}
\end{table}

\begin{table*}[t]
\begin{center}
{\def\arraystretch{1.2}\tabcolsep=5pt
\begin{tabular}{@{}lllllllllllllllll@{}}
\toprule
\multicolumn{1}{c}{\multirow{2}{*}{Method}} & \multicolumn{4}{l}{Surgery} & \multicolumn{4}{l}{Dough-Rolling} & \multicolumn{4}{l}{Drawing} & \multicolumn{4}{l}{Chopstick-Using} \\
\cmidrule(l){2-17}
\multicolumn{1}{c}{} & S & T & TS & & S & T & TS & & S & T & TS & & S & T & TS \\ 
\cmidrule{1-17}
Siamese TSN with $L_{rank1}$ & 64.7 & 72.8 & 69.1 & & 77.6 & 79.4 & 78.5 &  &75.6 & 77.4 & 78.0 &  &67.2 &  67.9 & 68.8\\
Siamese TSN with $L_{rank2}$ & 64.4 & \textbf{73.3} & 69.0 & & 79.1 & \textbf{80.4} & 78.5 & & 74.9 & 81.8 & 79.1 & & 67.2 & 69.9 & 68.8\\
Siamese TSN with $L_{rank3}$ & \textbf{66.4} & 72.5 & \textbf{70.2} & & \textbf{79.5} & 79.5& \textbf{79.4}& & \textbf{77.6} & \textbf{82.7}& \textbf{83.2}&  & \textbf{70.8} & \textbf{70.6} & \textbf{71.5} \\

\end{tabular}
}
\end{center}
\vspace{-1mm}
\caption{Results of 4-fold cross validation on all datasets, for our proposed method with each of our proposed loss functions. For all datasets $L_{rank3}$ outperforms original loss $L_{rank1}$. S=Spatial, T=Temporal, TS=Two-Stream} 
\label{tab:base}
\vspace{-1mm}
\end{table*}

As well as the \textit{consistent pairs} we use in $L_{rank}$ we also require similarly ranked pairs for $L_{sim}$ (Eq.~\ref{eq:l_sim}). 
These are not all the \textit{inconsistent pairs}, as those may be noisy.
%It is important to note that we cannot take all the pairs which are not \textit{consistent pairs} are the \textit{AMT} annotations will contain noise, meaning not all \textit{inconsistent pairs} should be considered similar. 
%Therefore, we need a secondary measure to ensure the pairs selected from the set of \textit{inconsistent pairs} are of a similar skill level. We do this 
We select similar pairs for training using the directed graph of all pairs introduced above. We define separation between a pair of videos to be the difference in the length of the longest walk from any source node in the graph. We consider pairs in the set of \textit{inconsistent} pairs with a separation of 0 or 1 as our set of \textit{similar pairs}. %The amount of similar pairs we retrieve from each dataset is shown in Table~\ref{tab:pairs}.

Table~\ref{tab:pairs} presents statistics on these \textit{consistent} and \textit{similar} pairs. 
%These consistent pairs are the pairs we use for evaluation and in the training process with the pairwise margin loss.
%\DimaN{[again, this last sentence is no longer true, right?]}
\textbf{Surgery} has a high number of consistent pairs ($>95$\%). 
The pairs in this dataset come from the scores of a single expert, available with the JIGSAWS dataset, therefore pairs are only excluded when two videos have the same score.
For the other tasks, we use the judgments from multiple AMT workers. 
\textbf{Dough-Rolling} has the lowest percentage of \textit{consistent }pairs, as many were considered comparable in skill by human annotators. 
This is likely due to the nature of the task, thus many subjects do manage a similar level of performance.
For \textbf{Drawing} and \textbf{Chopstick-Using}, the number of consistent pairs is $60-70$\%.

%\DimaN{[I didn't understand the next paragraph, which I assume is important, so let's discuss (keep comment until sorted). Do we also need to report these new numbers in Table 1? to know how much we use pair dataset]\Hazel{Any clearer?}}

%With our new pairwise similarity loss we add pairs which should be ranked similarly into our learning process. We can't simply take all pairs for which \textit{AMT} workers disagree on as annotations on \textit{AMT} can contain noise, this is why we ensure agreement to find our consistent pairs. Therefore, we need a secondary measure to ensure the pairs we take from the inconsistent pairs should be similarly ranked. 
%We do this with the directed graph of all pairs mentioned previously. We extract an auxiliary ranking for each video by giving each video a rank of the length of the longest walk from the source node(s) in the graph. Pairs with a ranking separation of 1 or less are than used in training as similar pairs. 

%All four tasks with their annotations and consistent pairs are combined into a public dataset for skill determination in video available at: \textit{Link available with publication}

\section{Experiments}
\label{sec:experiments}
For all datasets, we use a four-fold cross validation to report results. For each fold, the pairs between three quarters of the videos are used in training, and we then test on all remaining pairs. This includes pairs where neither video has been used in a pair for training as well as pairs where one video has been used in training within a different pairing.

\subsection{Implementation Details}
To extract the optical flow frames for the temporal network we use the $TV-L^{1}$ algorithm~\cite{zach2007duality}. 
We use mini-batch stochastic gradient descent with a batch size of 128 and a momentum of 0.9. Both the spatial and temporal network use the AlexNet~\cite{krizhevsky2012imagenet} architecture as we found this gave better results with a shorter training time than the BN-Inception~\cite{ioffe2015batch} network original used in TSN. Both sides of each Siamese network are initialized with network weights from pre-trained ImageNet models. %\DimaN{[do we want to say that we tried VGG?]}\Hazel{Will add in a comment about BN-Inception once we've got the final Surgery result.}
In the spatial network the learning rate begins as 1E-3 and decreases by a factor of 10 every 1.5K iterations, with the learning process finishing after 3.5K iterations. The temporal network's learning rate is initialized as 5E{-3}, decreasing by a factor of 10 after 10K iterations and after 16K iterations, with learning ending after 18K iterations. 
We set $\beta$ (Eq.~\ref{eq:finalLoss}) to 0.5 in all experimental results after initial assessment.

To avoid over-fitting, we use the same data augmentation techniques as Wang \etal~\cite{wang2016temporal}, namely horizontal flipping, corner cropping and scale jittering on the 340x256 pixels RGB and optical flow images. The cropped regions are 224x224 pixels for network training.
We also use dropout layers with the fully connected layers, ratios used are 0.5 for both streams.

\subsection{Evaluation Metric}
To evaluate our method, we use pairwise precision on the rankings produced by each testing fold. Pairwise precision is defined as the \textit{percentage of correctly ordered pairs in a ranking}. We say a pair is correctly ordered if for a pair $(p_i, p_j)$ where $E(p_i, p_j) = 1$ in the ground truth, the method outputs $f(p_i) > f(p_j)$.

\begin{figure*}[t]
\begin{minipage}{0.28\textwidth}
\includegraphics[width=1\linewidth]{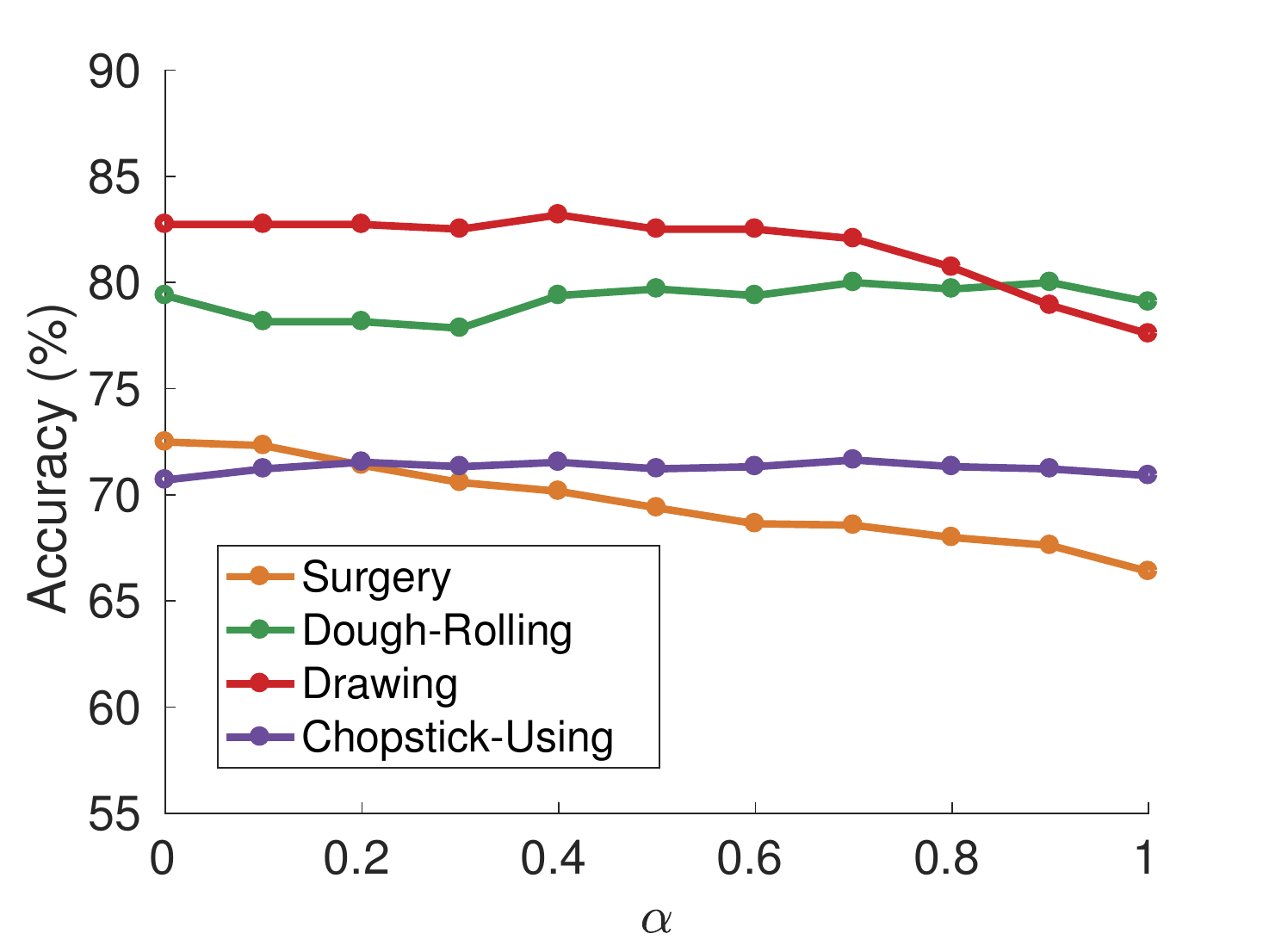}
	\caption{The accuracy for each dataset with different $\alpha$ values. The method is resilient to the parameter value chosen.}
	\label{fig:alpha}
\end{minipage}
\hspace{0.03\textwidth}
\begin{minipage}{0.65\textwidth}
			\includegraphics[width=0.32\linewidth]{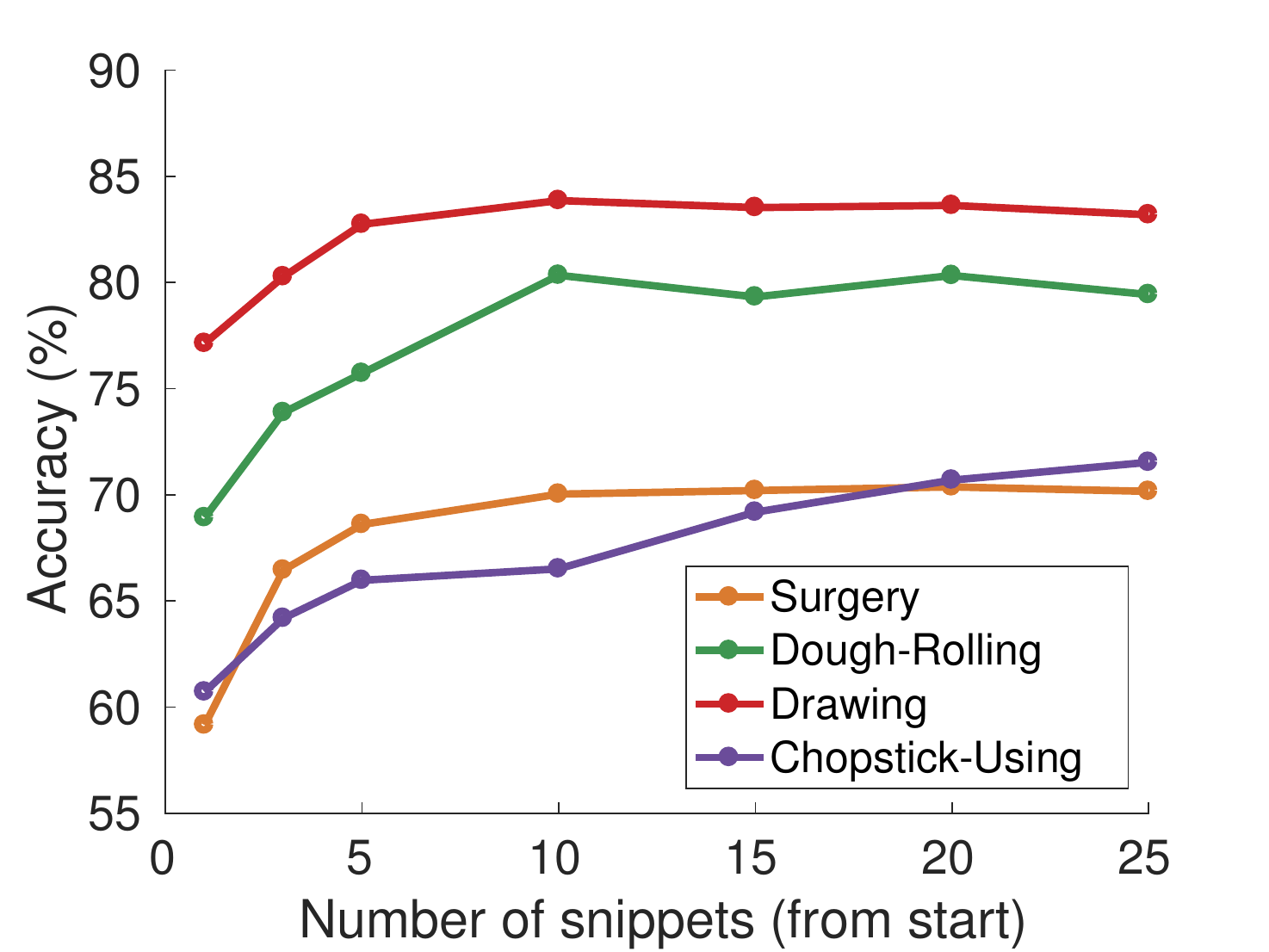}
            \includegraphics[width=0.32\linewidth]{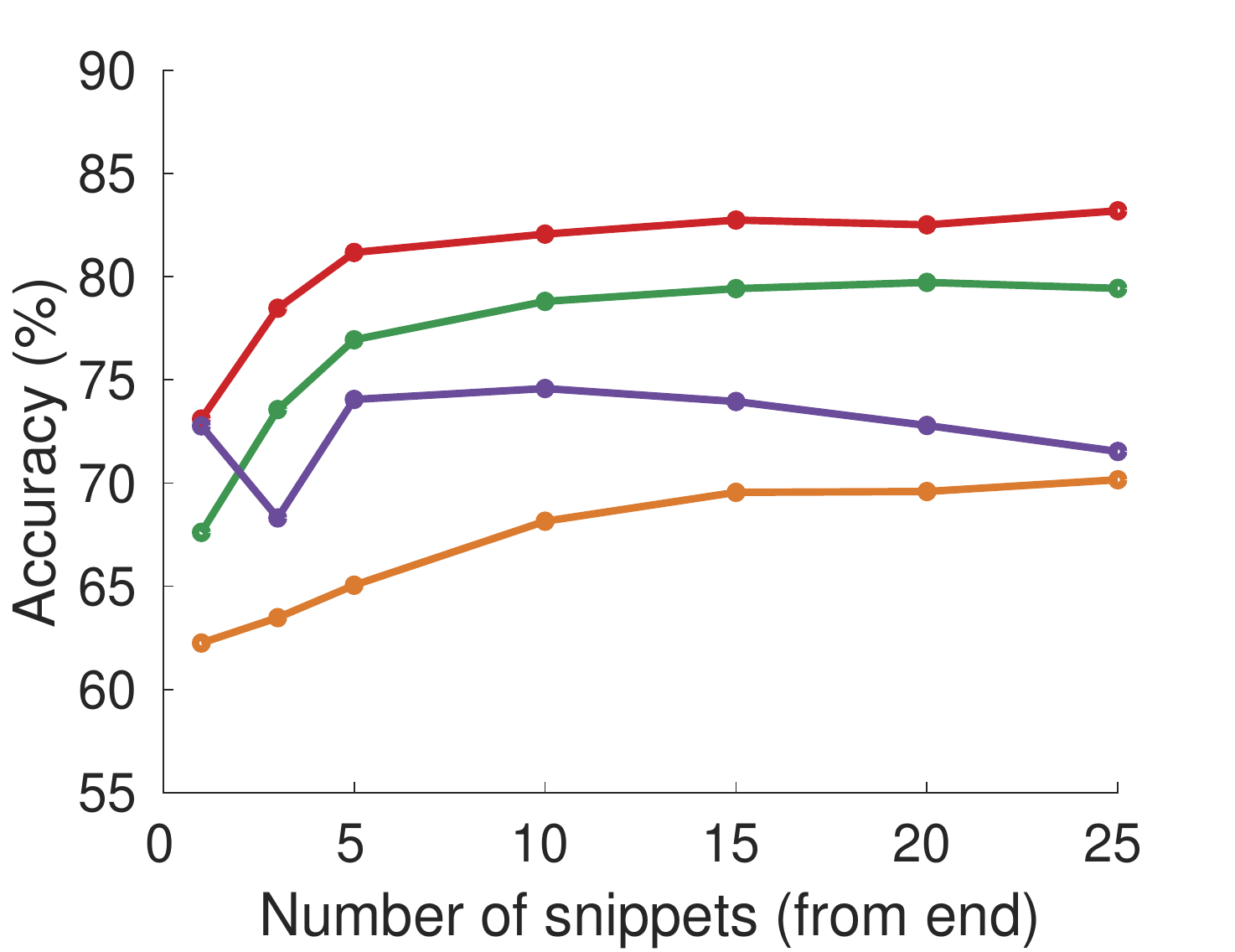}
            \includegraphics[width=0.32\linewidth]{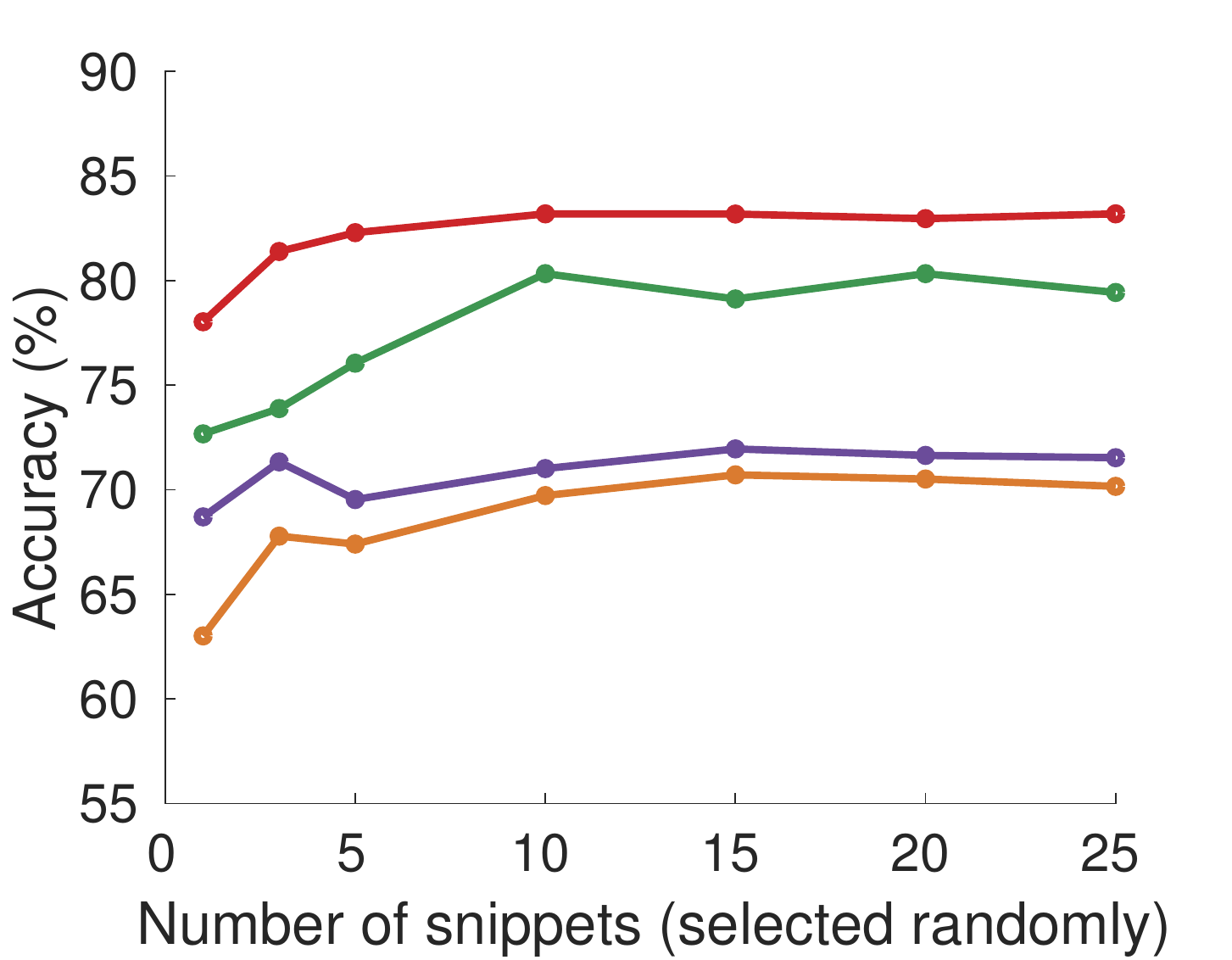}
	\caption{The accuracy achieved when adding snippets in testing, from the start (left), end (middle) and randomly (right).}
	\label{fig:frame_test}
\end{minipage}
\vspace{-3mm}
\end{figure*}

\subsection{Results}
%Baselines
%\Hazel{Rewrote}
In Table~\ref{tab:base} we show our results from four-fold cross-validation on each of the four datasets with each loss function. We report results with  $\sigma = 25$ as in~\cite{wang2016temporal}, and for $\alpha=0.4$ (Eq.~\ref{eq:alpha}) as in ~\cite{simonyan2014two,wang2016temporal}. Below we test the sensitivity of these results to the values of $\alpha$. From this Table~\ref{tab:base}, we can see that our proposed loss function $L_{rank3}$ outperforms the standard margin loss function $L_{rank1}$ on all combinations of modality and dataset. We also see an improvement from $L_{rank2}$ over all but the temporal result in Surgery and Dough-Rolling. This improvement is particularly noticeable in the two-stream results for Drawing (79.1\% to 83.2\%) and Chopstick-Using (68.8\% to 71.5\%). The inclusion of similar pairs with $L_{sim}$ in the training process has the largest impact on training on the spatial network, where the results for skill determination are generally weaker across tasks. $L_{sim}$ increases the spatial network performance %\DimaN{[increases this performance? not sure what that means]} 
resulting in larger improvements in the two-stream result. %\Hazel{Why?}

From the results in Table~\ref{tab:base} we can also conclude that the temporal features are in general more useful for determining skill for the presented tasks, with the temporal result outperforming the spatial result in all but the Chopstick-Using task. Although, we do manage to reduce this gap with our additional loss $L_{sim}$ %\DimaN{[now there's sim3!]} 
from Section~\ref{sec:training_pairwise}. This implies the motions performed are more important for determining skill than the current state or appearance of the task (captured in the spatial stream). We note that the largest difference between the two streams is in the Surgery tasks. This is because these tasks require quick smooth motions, putting minimal stress on the surrounding areas. Hence, while the end result of each stage is visually similar, the motions affect the scoring significantly.

%We also test the data augmentation technique described in~\ref{sec:dataaug}. 
%From Table~\ref{tab:base}, we see that our data augmentation for skill videos consistently improves upon the basic Siamese TSN. This is particularly true for the Dough-Rolling (75.4\% to 78.2\%) and Drawing tasks (77.4\% to 82.1\%), as these datasets are smaller, resulting in fewer training pairs. We can also see that our pairwise similarity loss consistently improves results. \Hazel{TODO: Add in more when have temporal}

\medskip
\noindent\textbf{Fusion Parameter.}
We assess the sensitivity of our results to the late fusion weighting $\alpha$ in Equation~\ref{eq:alpha}. We test $\alpha$ values from 0 to 1 at intervals of 0.1 for all datasets, as shown in Figure~\ref{fig:alpha}. 
%From Figure~\ref{fig:alpha} we confirm that, 
For the majority of tasks, the combination of temporal and spatial modalities is useful, except the Surgery task which peaks at $\alpha=0$, i.e. no information from the spatial network is included. 
%This highlights the difference between what it means to be skilled in different tasks. In the Surgery tasks the quality of motion and size of the motion used are much more important than in the Dough-Rolling task, where more can be seen from the result of the task. 
%Despite this, 
All tasks benefit from the contribution of the temporal network, albeit to different degrees.
%, demonstrating the importance of the temporal network. 
We also observe that the method is highly resilient to the value of $\alpha$, particularly for the Chopstick-Using and Dough-Rolling tasks. %\Hazel{It was more sensitive to change in the previous iteration, might be worth mention this?}

\medskip
\noindent\textbf{Number of Snippets in Testing.}
The results above are reported for $\sigma=25$ (in-line with previous methods), where the snippets are sampled uniformly from the whole video. However, it is interesting to examine how much of the video is needed at test time to gain an accurate evaluation of the skill displayed in a video. We test our evaluation on a varying number of consecutive snippets from the start and end of the video, as well as randomly. Results are shown in Fig.~\ref{fig:frame_test}. 

%\medskip
%\noindent\textbf{Number of Snippets in Testing.}
%Previous results are reported for $\sigma = 25$ snippets, similar to~\cite{wang2016temporal}. We assess the effect of $\sigma$ on performance.
%When $\sigma = 1$, we select one snippet at the centre of the video, for $\sigma = 3$, we select a snippet at the start, one at the centre and one at the end. Snippets are uniformly sampled for $\sigma = 5, 10, 20, 25, 35, 50$.
%Results are shown in Figure~\ref{fig:frame_test}.

%\begin{figure*}[t]
%	\begin{center}
%			\includegraphics[width=\textwidth]{Result_Graphs/crop_agg}
%	\end{center}
%	\caption{The accuracy of each dataset using four functions for segmental consensus, with $\alpha=0.4$ and $\sigma=25$.}
%	\label{fig:consensus}
%\end{figure*}

%From Figure~\ref{fig:frame_test} we see little improvement after $\sigma = 25$ snippets. We also note reasonable accuracy is achieved with only 5 snippets for all datasets. Interestingly, for Drawing, accuracy is the highest for $\sigma = 3$, with the lowest at $\sigma = 1$. This indicates discriminative information is contained in the snippets from the first and last parts of the video, showing the initiation of drawing as well as its completion. This is in contrast with the Surgery tasks, as mistakes will have been corrected by the last snippet meaning little useful information is contained there. 

From Figure~\ref{fig:frame_test} we can see that good accuracy can be obtained after only seeing a portion of the video, however a single frame of the video is insufficient to measure skill, even if this single frame contains the end result of the task, and accuracy improves as further snippets are viewed. Interestingly, accuracy converges as the number of snippets continues to increase, %the amount of video seen at which the accuracy no longer changes is different 
for the various tasks and snippet sampling approaches. 
For instance, the Surgery task achieves near peak accuracy with the first 20\% of the testing snippets, while the last 20\% appear redundant. This difference is intuitive as the start of the Surgery task is more challenging, while the repetitive nature of the task allows novice participants to improve by the end.

\subsection{Baselines}
As there are no generic existing methods for ranking skill nor performing skill determination for non-surgical tasks, we use existing ranking methods developed for other applications. Our first baseline uses RankSVM~\cite{joachims2002optimizing}, commonly used in ranking problems~\cite{parikh2011relative}. We perform four-fold cross validation on RankSVM with features extracted from Alex{N}et trained on ImageNet~\cite{krizhevsky2012imagenet} and C3D trained on Sports1M~\cite{tran2015learning}. These two results are then combined with late fusion of $\alpha=0.4$. We use the same implementation as in~\cite{joachims2006training} which can be found on the authors webpage.
 
%Importantly, no available baseline on skill determination from video has been published on the Surgery dataset to date - published results report gesture classification solely~\cite{ahmidi2017dataset}. Kinematic data from the Surgery dataset has been used~\cite{zia2017automated} but, due to the different data type, these methods are not comparable to ours.

\begin{figure*}[t]
	\begin{center}
			\includegraphics[width=\textwidth]{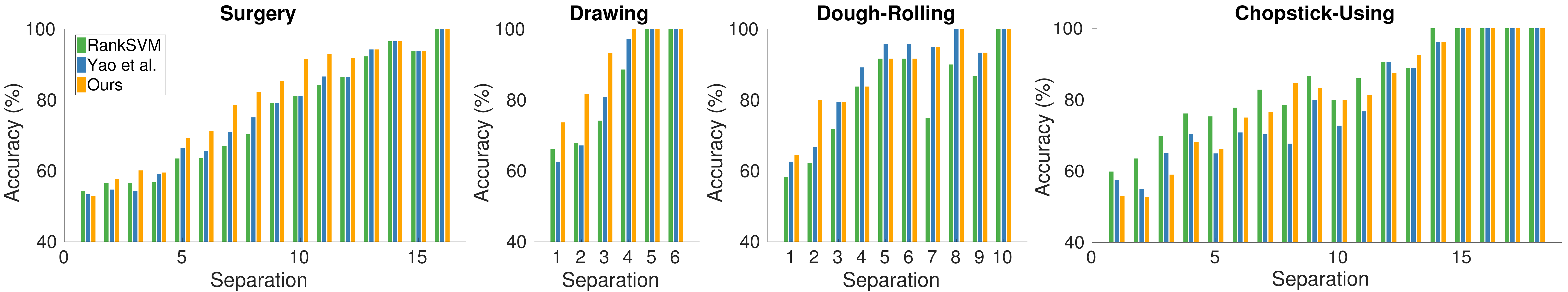}
	\end{center}
    \vspace{-2mm}
	\caption{The accuracy of each ordered pair by separation between videos in a pair for each task, in the baselines and our method. The accuracy consistently increases as tested pairs are further in the ground-truth ranking for all datasets.}
	\label{fig:separation}
    \vspace{-2mm}
\end{figure*}

%\begin{figure*}[t]
%	\begin{center}
%		\includegraphics[width=0.49\textwidth]{Rankings/sonic_ranking_compressed}
        %\includegraphics[width=0.49\textwidth]{Rankings/hand_ranking_new}
		%\includegraphics[width=0.49\textwidth]{Rankings/sonic_ranking2}
%        \includegraphics[width=0.49\textwidth]{Rankings/hand_ranking2_new_compressed}
% 		\includegraphics[width=\textwidth]{Rankings/cmu_rank2_compressed}
%	\end{center}
%    \vspace{-3mm}+
%	\caption{Ranking examples from the Drawing and Dough-Rolling tasks from multiple folds.}
%	\label{fig:rankingeg}

%\end{figure*}

\begin{table}[t]
\begin{center}
{\def\arraystretch{1.2}\tabcolsep=5pt
\begin{tabular}{@{}lllll@{}}
\toprule
\multicolumn{1}{c}{\multirow{2}{*}{Method}} & \multicolumn{1}{l}{\multirow{2}{*}{Surgery}} & \multicolumn{1}{l}{Dough-} & \multicolumn{1}{l}{\multirow{2}{*}{Drawing}} & \multicolumn{1}{l}{Chopstick-} \\
& & Rolling & & Using \\
\cmidrule{1-5}
RankSVM~\cite{joachims2006training} & 65.2  & 72.0 & 71.5 & \textbf{76.6}\\
Yao \etal~\cite{yao2016highlight} & 66.1 & 78.1 & 72.0 & 70.3\\ 
Ours&  \textbf{70.2} &  \textbf{79.4}& \textbf{83.2}&  71.5 \\
\end{tabular}
}
\end{center}
\vspace{-1mm}
\caption{Results of 4-fold cross validation on all datasets, for the baselines  our proposed method with $L_{rank3}$.}
\vspace{-2mm}
\label{tab:baselines}
\end{table}

The second baseline is Yao \etal~\cite{yao2016highlight} who originally performed deep ranking on video to determine highlight segments. They use pre-extracted features in a fully connected network to determine segments with the highest potential highlight score. The features used are from Alex{N}et and C3D which are then passed separately into a network with architecture: $F$1000-$F$512-$F$256-$F$128-$F$64-$F$1 using the same margin loss in Eq.~\ref{eq:loss}. The results from each network are then fused using late fusion with $\alpha=0.4$.  Although this method was originally developed for a different purpose - binary highlight detection -
%there is nothing specific to highlight detection in the ranking process. Therefore 
it 
%could be applicable as 
uses a general method of ranking video and is thus used here for comparison.%determining skill in many tasks than previous domain-specific skill determination work, so it is used for comparison here.
%\DimaN{[this second baseline requires full rewrite. What features? how many fully connected layers? what size? same architecture as paper]}\Hazel{Is this section better now}
%we use the average performance of random rankings of the videos in each task as a baseline, i.e. 50\% accuracy\footnote{Each pair in a random ranking has two possibilities: to be correctly or incorrectly ordered. For each possible ranking with performance $x$, there exists the reverse ranking with performance $1-x$. Hence baseline of 50\%.}.

It is important to note, the only dataset for which skill evaluation has previously been considered is the Surgery dataset, though as a regression problem to expert scores. This approach is not applicable to daily tasks where obtaining objective scoring is much harder than pairwise ranking. Published results on the Surgery dataset either report Expert/Novice classification~\cite{ahmidi2017dataset} or use only the kinematic data~\cite{zia2017automated} and are therefore not comparable to ours.

Comparative results are available in 
%The comparison between our method and the baselines can be seen in 
Table~\ref{tab:baselines}. Our method outperforms both baselines on three of the four tasks.
%in all tasks except Chopsticks, where the 
RankSVM performs best on Chopstick-Using.
%out of the three.
%\DimaN{[do we have any insight why???]}\Hazel{Unfortunately not}
The improvement with our method is most significant in the Drawing task with an improvement of 11.2\%. 

To study where the difference in performance lies, we show the accuracy for each level of separation between pairs of each dataset in Figure~\ref{fig:separation}. Assume we have consistent annotation pairings resulting in the partial ranking ${p_i < p_{i+1} < ... < p_{i+n} < p_j}$, then we define separation between $p_i$ and $p_j$ as $n+1$. It is more important that pairs with high separation be correctly ordered than pairs close together in the ranking.
Figure~\ref{fig:separation} shows that the significant improvement of our method, over the baselines, in Surgery and Drawing is at the mid-level of separation. Although all methods approach the 100\% accuracy for the most separated pairs, our method approaches this much faster. Alternatively, in the Chopstick-Using task, the only task for which we perform below one baseline, we have comparative performance in the mid and high separation compared to RankSVM, only falling below for nearby pairs.

%While Table~\ref{tab:base} reports results for pairwise comparisons, we also wish to assess which pairs are being correctly ordered. 
%Assume we have consistent annotation pairings resulting in the partial ranking ${p_i < p_{i+1} < ... < p_{i+n} < p_j}$. It is more important that the pair $(p_i, p_j)$ is correctly ordered by our method, as opposed to the pair $(p_i, p_{i+1})$. We test this by reporting accuracy as separation between pairs in the ranking increases. For the partial ranking above, we define the separation of the pair $(p_i,p_{i+1})$ to be $1$ and of the pair $(p_i,p_j)$ to be $n+1$. 

%From Figure~\ref{fig:separation} we observe, that as the separation between videos increases so does the accuracy of the correctly ordered pairs, reaching 100\% for the furthest pairs.

%\DimaN{[I don't think that last sentence was necessary. Let your results talk on your behalf]}
%We can therefore conclude that our method is the most suitable for skill determination in a wide variety of tasks.\textbf\textbf{}{} 

\begin{figure}[t]
\begin{center}
\vspace{-2mm}
\includegraphics[width=0.49\textwidth]{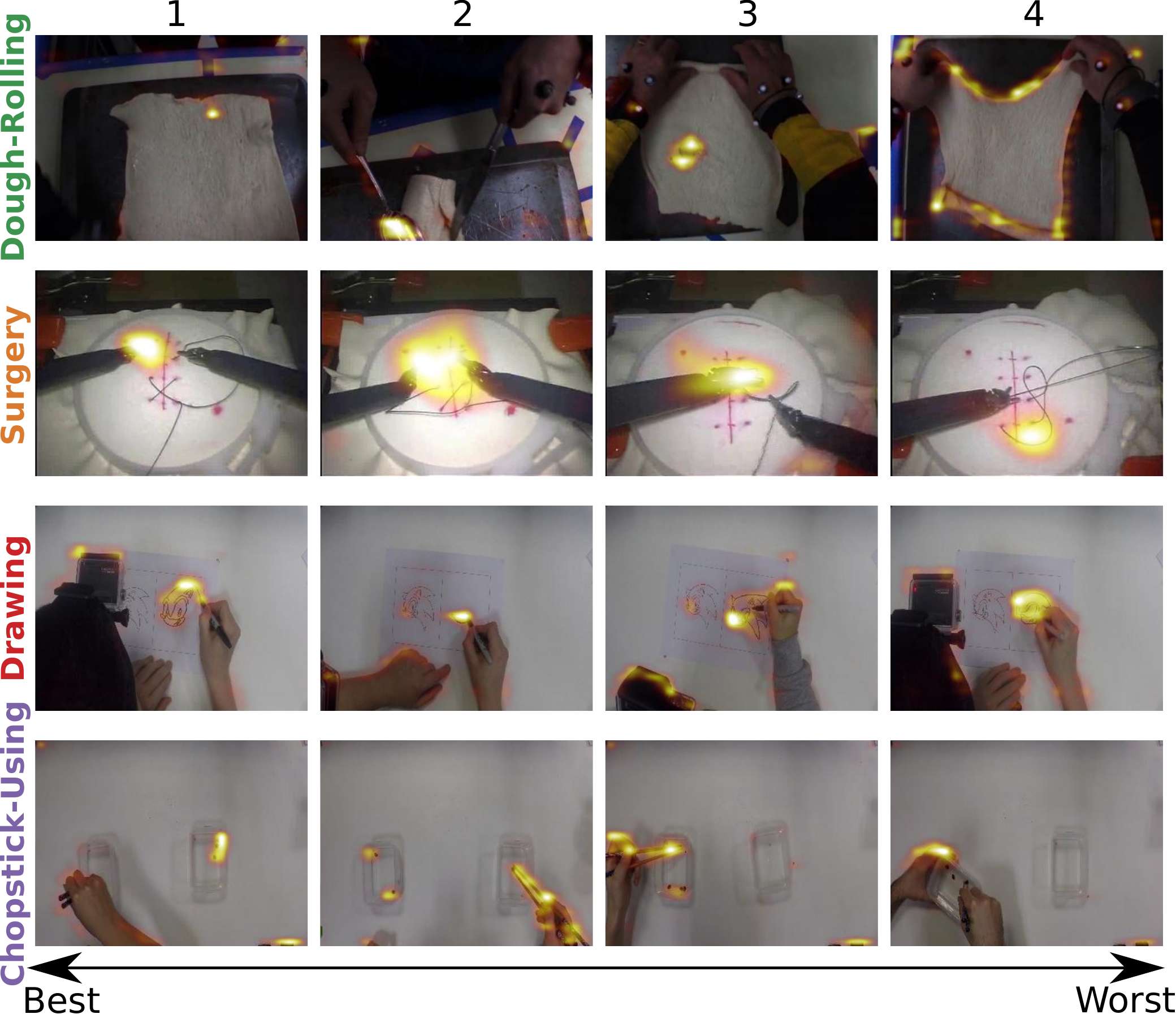}
\end{center}
\vspace{-0.8em}
\caption{Spatial activations for sample frames at varying ranks.}
\vspace{-1.1em}
\label{fig:vis}
\end{figure}

\subsection{Visualizing Performance Ranking}

A key difficultly of skill determination is capturing the nuance of the tasks in the learned model. In Fig.~\ref{fig:vis} we visualize the top-down attention of the spatial CNN on example rankings for three datasets using~\cite{willprice} based on~\cite{zhang2016top}. For each dataset, we show frame-level spatial activations on four videos with varying levels of skill (best$\rightarrow$worst). 

From Fig.\ref{fig:vis} we can see that the trained model is picking details that correspond to what a human would attend to. In Dough-Rolling high activations occur on holes in the dough (1, 3), curved or rolled edges (4) and when using a spoon (2). High activations occur in Surgery when strain is put on the material (1, 2), with abnormal needle passes (3) and when there is loose stitching (4). In Drawing, the model attends to specific parts of the sketch such as the head and mouth. The high activations in the Chopstick-Using task occur on the hand position (3,4), chopstick position (2) and the bean locations (1,2,3).
Further qualitative results are shown in the supplementary video.
%Fig.~\ref{fig:vis} also demonstrates the necessity of using video to rank skill as we see activations during the video e.g. loose knots and patches in dough that are later fixed in the video towards a `perfect' final frame.

%Clearly a single frame from the video is not sufficient to visualize skill or ranking. Further results displaying the activations from the temporal stream are thus available in the supplementary material on the authors' webpage.
%For Drawing, the results from the higher ranked videos have a better resemblance to the reference image. 
%In the Sonic-Drawing rankings we see visible improvement in the shape and expression of Sonic from left to right. In the Hand-Drawing task the method manages to lowly rank the videos in which the participants draw the perspective incorrectly or add little detail.

%In Figure~\ref{fig:rankingeg}, we also demonstrate an increase in skill from lowest to highest ranked videos for the Dough-Rolling task. For example, the second image shows dough with a badly patched hole. The third image shows an improvement, with the dough containing only several small holes. Again the fourth image demonstrates more skill, as holes have been patched less obviously. 

%Clearly, a single frame from the video is not sufficient to visualize skill or ranking. Further results are thus available in the supplementary video. %\DimaN{[You know, saying the previous sentence: video on the authors' page disqualifies the whole paper as non-anonymous, so you should be careful]}
\section{Conclusion}

In this paper we have presented a method to rank videos based on the skill that subjects demonstrate. Particularly, we have proposed a pairwise deep ranking model which utilizes both spatial and temporal streams in combination with a novel loss to determine and rank skill. We have tested this method on four separate datasets, two newly created, and show that our method outperforms the baseline on three out of four datasets, with all tasks achieving over 70\% accuracy. Furthermore, we have explored where the performance increase lies and examined our method's resistance to changes in parameters. Qualitative figures demonstrate the approach's ability to learn tasks' nuances, while using a general, task-independent, method.

We see our work as a promising step toward the automated and objective organization of \textit{how-to} video collections and as a framework to motivate more work in skill determination from video. Further work involves exploring mid-level fusion between the two streams of the network, as well as testing on additional and across datasets and tasks.

\noindent \textbf{Acknowledgements:} Access to EPIC-Skills 2018 dataset and annotations available from authors' webpages. Supported by  an EPSRC DTP and EPSRC GLANCE (EP/N013964/1).

{\small
\bibliographystyle{ieee}
\bibliography{egbib}
}

\end{document}